\documentclass[letterpaper]{article} 
\usepackage{aaai2027}  
\usepackage[hyphens]{url}  
\usepackage{graphicx} 
\usepackage{amsmath}
\usepackage{natbib}  
\usepackage{caption} 
\usepackage{algorithm}
\usepackage{algorithmic}

\usepackage{newfloat}
\usepackage{listings}
\DeclareCaptionStyle{ruled}{labelfont=normalfont,labelsep=colon,strut=off} 
\floatstyle{ruled}
\newfloat{listing}{tb}{lst}{}
\floatname{listing}{Listing}

\usepackage{booktabs}

\title{D²-4DGS: Dual-Depth Guided Sparse-Camera 4D Gaussian Splatting}
\author{    Jijian Zhao}\affiliations{    Huazhong University of Science and Technology\\    jijianzhao28@gmail.com}

\usepackage[table]{xcolor}
\usepackage{booktabs}

\DeclareRobustCommand{\legendbest}[1]{%
  \begingroup\setlength{\fboxsep}{1pt}\colorbox{red!25}{\strut\textbf{#1}}\endgroup}
\DeclareRobustCommand{\legendsecond}[1]{%
  \begingroup\setlength{\fboxsep}{1pt}\colorbox{orange!25}{\strut #1}\endgroup}
\DeclareRobustCommand{\legendthird}[1]{%
  \begingroup\setlength{\fboxsep}{1pt}\colorbox{yellow!25}{\strut #1}\endgroup}

\newcommand{\capbest}[1]{\cellcolor{red!25}\textbf{#1}}
\newcommand{\capsecond}[1]{\cellcolor{orange!25}#1}
\newcommand{\capthird}[1]{\cellcolor{yellow!25}#1}

\begin{document}

\maketitle

\begin{abstract}
Dynamic 4D Gaussian Splatting has emerged as an efficient representation for dynamic novel view synthesis through explicit scene modeling and real-time rendering. However, existing methods typically require dense multi-view videos for sufficient geometric constraints, making capture expensive and limiting sparse-camera deployment. Reducing input views lowers acquisition cost but weakens geometry supervision, often causing missing structures and floating Gaussians. Depth priors provide geometric cues, yet no single source offers both dense coverage and reliable geometry. Monocular depth provides dense structure but is scale-ambiguous and locally biased, whereas multi-view geometric depth provides incomplete anchors consistent with the reconstruction coordinate system. To exploit their complementarity, we propose D$^2$-4DGS, a sparse-camera dynamic 4D Gaussian Splatting framework guided by dual-source depth priors. We align monocular estimates with valid multi-view geometric depths and verify their consistency to identify reliable geometric anchors. These verified anchors support consistency-aware pruning and depth supervision, while verified geometric depths and aligned mono-only estimates provide candidate geometry for densification in under-reconstructed regions. Finally, RGB-D joint optimization improves appearance fidelity and geometric consistency under sparse-view supervision. Across all nine dataset--view settings, D$^2$-4DGS achieves the highest PSNR, improving by 1.33 dB on average over the best competing method in each setting.
\end{abstract}

\begin{figure}[!t]
    \centering
    \includegraphics[width=0.98\columnwidth]{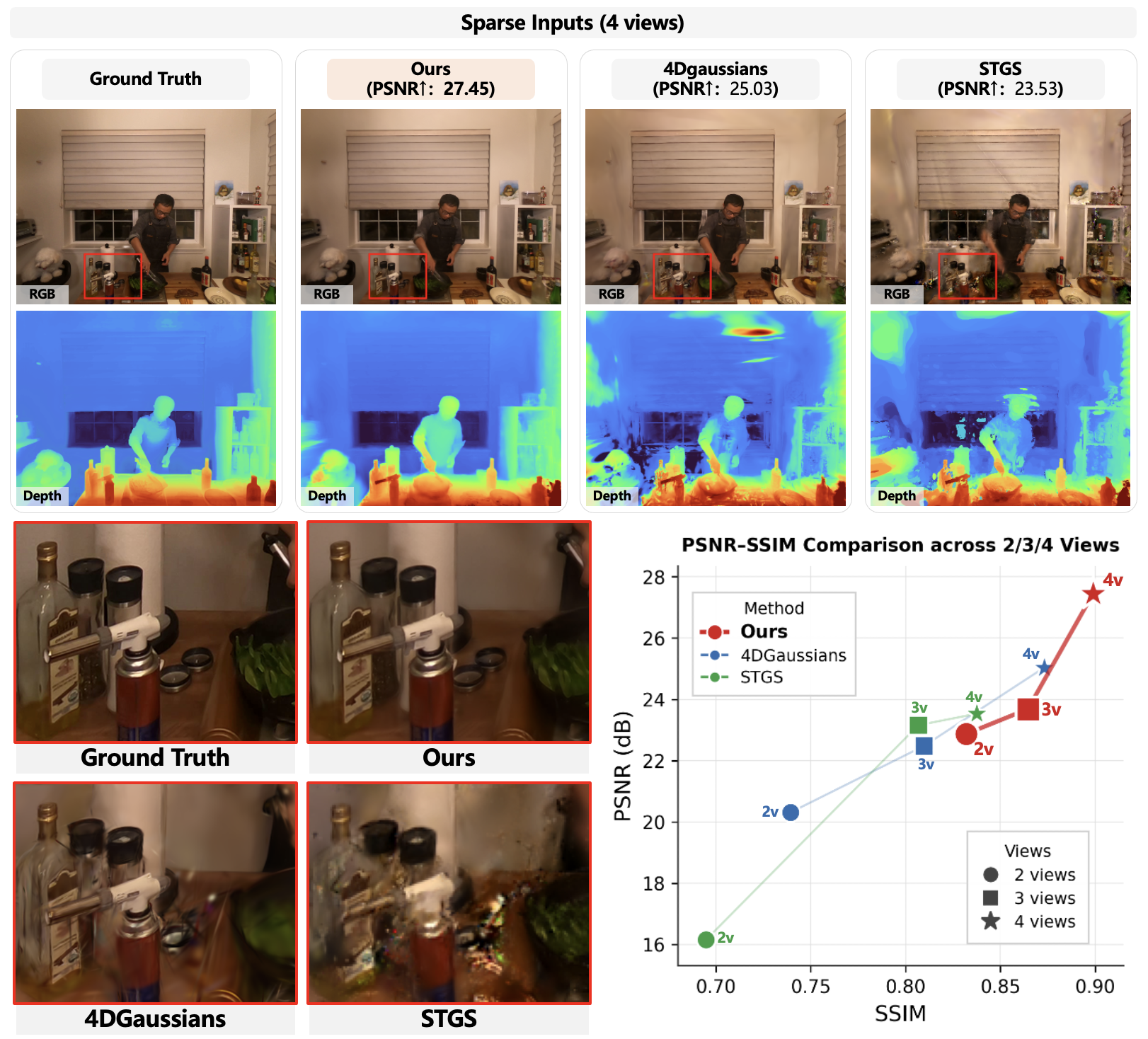}
    \caption{Sparse-camera dynamic reconstruction. Top: with four input views,
    D$^2$-4DGS produces sharper RGB renderings and more coherent depth maps
    than 4DGaussians and STGS. Bottom left: zoomed crops show that our method
    better preserves fine structures while reducing local distortions and
    floating artifacts. Bottom right: D$^2$-4DGS consistently achieves higher
    PSNR and SSIM under 2/3/4-view settings.}
    \label{fig:teaser}
\end{figure}

\section{Introduction}

Dynamic novel view synthesis reconstructs time-varying scenes and renders photorealistic unseen views, supporting immersive communication, virtual reality, digital humans, and robotic perception. 3D Gaussian Splatting (3DGS) offers explicit representation, differentiable rasterization, fast optimization, and real-time rendering \cite{kerbl20233d}. Recent 4D Gaussian methods extend 3DGS with temporal features, motion fields, or spacetime representations \cite{wu20244d,li2024spacetime}, retaining high visual fidelity and rendering efficiency.

However, most dynamic 4D Gaussian methods rely on dense multi-view videos. Their cross-view constraints facilitate Gaussian optimization and structure recovery, but require synchronized cameras, accurate calibration, and controlled acquisition. Using a few cameras reduces capture cost but weakens geometry supervision. Sparse observations and dynamic occlusion, motion, and deformation leave regions under-constrained, often producing inaccurate depth, missing structures and floating Gaussians.

Depth priors provide additional geometric cues for sparse-camera reconstruction, yet no single source offers both dense coverage and reliable geometry. Monocular depth provides dense scene structure but is scale-ambiguous and locally biased \cite{yang2024depth}, whereas multi-view geometric depth provides anchors consistent with the reconstruction coordinate system but remains incomplete under occlusion, weak texture, and limited baselines \cite{schonberger2016structure,schonberger2016pixelwise}. These two sources are complementary, but their predictions must be aligned and checked for consistency before providing reliable guidance.

To exploit this complementarity, we propose D$^2$-4DGS, a sparse-camera dynamic 4D Gaussian Splatting framework guided by dual-source depth priors. We align monocular estimates with valid multi-view geometric depths and retain consistent regions as reliable geometric anchors. These anchors support consistency-aware pruning and provide depth constraints during RGB-D optimization. For under-reconstructed regions, verified geometric depths and aligned mono-only estimates provide candidate geometry for Gaussian densification. RGB-D joint optimization then refines the appearance and geometry of the updated representation.

Our main contributions are summarized as follows.
\begin{itemize}
    \item We propose a sparse-camera dynamic 4D Gaussian Splatting framework, D$^2$-4DGS, which integrates dual-source depth priors into Gaussian structure optimization for high-quality dynamic novel view synthesis from limited camera inputs.
    \item We convert complementary depth evidence into explicit Gaussian structure updates by inserting primitives in under-reconstructed regions and jointly using verified depth inconsistency and visibility-aware opacity contribution to remove persistently unstable Gaussians.
    \item Extensive experiments on three dynamic datasets under 2/3/4-view settings show that D$^2$-4DGS achieves the highest PSNR across all nine dataset--view settings, with an average gain of 1.33 dB over the best competing method in each setting.
\end{itemize}

\begin{figure*}[t]
    \centering
    \includegraphics[width=\textwidth]{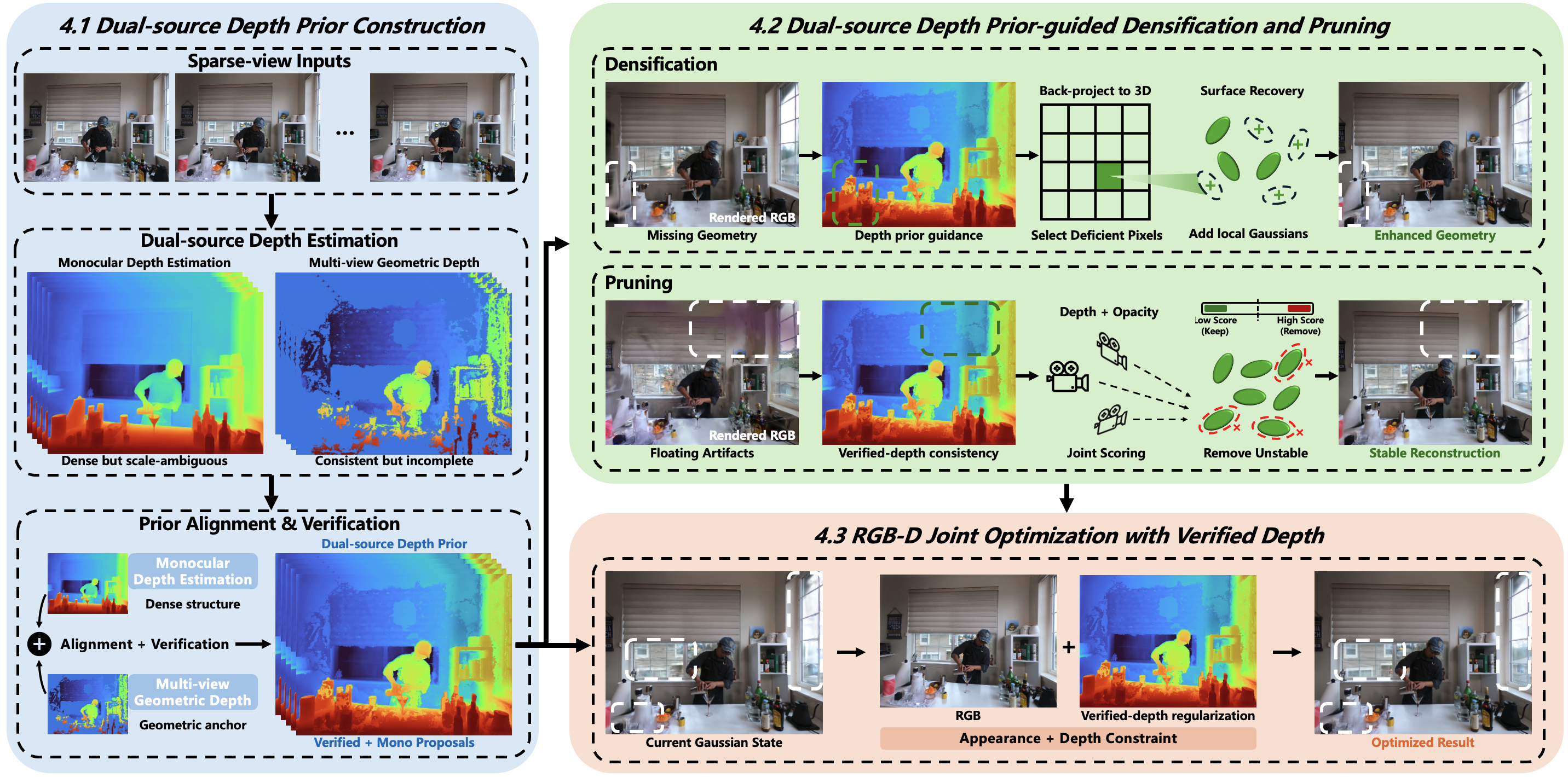}
    \caption{Framework overview. D$^2$-4DGS first aligns dense but scale-ambiguous monocular depth with geometrically consistent but incomplete multi-view depth from synchronized sparse views and retains their agreement regions as verified geometric anchors. These anchors guide pruning and depth supervision, while verified geometric depths and aligned mono-only estimates provide candidate geometry for local densification in under-reconstructed visible regions. Persistently unstable primitives are removed using contribution-weighted depth inconsistency and visibility-aware opacity contribution. Finally, RGB reconstruction and verified depth supervision jointly refine the updated 4D Gaussian representation with sparse-camera inputs.}
    \label{fig:pipeline}
\end{figure*}

\section{Related Work}

\subsection{Dynamic 4D Gaussian Splatting}
3DGS supports novel view synthesis with explicit primitives, differentiable rasterization, fast optimization, and real-time rendering \cite{kerbl20233d}. Dynamic extensions model temporal variation using deformation fields \cite{yang2024deformable}, temporal Gaussian features \cite{wu20244d}, persistent trajectories \cite{luiten2024dynamic}, spacetime features \cite{li2024spacetime}, or spatiotemporal primitives \cite{yang2024real}. Rotation-aware modeling \cite{duan20244d}, motion-aware representations \cite{lin2024gaussian}, and temporal regularization \cite{guo2024motion} further improve reconstruction. Yet these methods generally rely on dense multi-view videos, whereas sparse cameras weaken geometry and motion constraints, causing inaccurate depth, missing structures, and floating artifacts.

\subsection{Sparse-view and Sparse-camera Gaussian Reconstruction}
Sparse-view Gaussian methods introduce geometric constraints. SparseGS uses geometry-aware regularization \cite{xiong2023sparsegs}, FSGS grows Gaussians \cite{zhu2024fsgs}, depth-regularized 3DGS introduces external depth \cite{chung2024depth}, CoR-GS co-regularizes appearance and geometry \cite{zhang2024cor}, and DNGaussian normalizes global-local depth \cite{li2024dngaussian}. Feed-forward alternatives include pixelSplat \cite{charatan2024pixelsplat}, MVSplat \cite{chen2024mvsplat}, and InstantSplat \cite{fan2024instantsplat}, while 4C4D strengthens geometry in four-camera dynamic reconstruction through neural opacity decay \cite{zhou20264c4d}. However, most methods target static scenes or feed-forward prediction. Sparse-camera dynamics additionally involve fragmented geometry, varying visibility, and occlusion, requiring structure-aware Gaussian addition and removal.

\subsection{Depth Priors for Reconstruction}
Depth priors supplement RGB observations. MiDaS \cite{ranftl2020towards}, DPT \cite{ranftl2021vision}, and Depth Anything \cite{yang2024depth} provide dense monocular estimates but remain scale-ambiguous. Multi-view geometry supplies geometric anchors consistent with the reconstruction coordinate system through COLMAP-based SfM \cite{schonberger2016structure} and PatchMatch MVS \cite{schonberger2016pixelwise}. MVSNet \cite{yao2018mvsnet}, CasMVSNet \cite{gu2020cascade}, and PatchmatchNet \cite{wang2021patchmatchnet} improve matching, while TransMVSNet \cite{ding2022transmvsnet} and MVSFormer \cite{cao2022mvsformer} introduce transformers. Under sparse dynamic inputs, multi-view depth remains incomplete because of limited baselines, occlusion, weak texture, and motion. Thus, neither source alone provides both dense coverage and geometrically reliable guidance, motivating their combination with explicit Gaussian structure optimization.

\section{Preliminary}

We use a representative 4D Gaussian formulation \cite{yang2024real} to introduce notation. In this formulation, a dynamic scene is modeled as a set of anisotropic Gaussian primitives in the spatiotemporal domain $(\mathbf{x},t)$. Each primitive is parameterized by a 4D mean $\mu=(\mu_x,\mu_t)$, where $\mu_x$ denotes its 3D spatial center, a 4D covariance matrix $\Sigma$, an opacity $\alpha \in [0,1]$, and appearance coefficients $c$. Similar to 3DGS, the covariance is commonly decomposed by a 4D rotation and diagonal scaling:
\begin{equation}
\hspace{-0.025em}\Sigma=R\mathrm{diag}(s^2)R^T,\; R\in SO(4),\; s=(s_x,s_y,s_z,s_t).
\end{equation}

To render a frame at timestamp $t$, each 4D Gaussian is sliced into a 3D Gaussian by conditioning on the temporal coordinate, yielding a time-dependent 3D mean $\mu_{3D}(t)$, a 3D covariance $\Sigma_{3D}$, and a temporal attenuation term $F(t) \in (0,1]$. The sliced density at time $t$ is
\begin{equation}
\mathcal{G}_{3D}(\mathbf{x},t)=F(t)\exp\left(
-\frac{1}{2}\Delta^T(t)\Sigma_{3D}^{-1}\Delta(t)
\right),
\end{equation}
where $\Delta(t)=\mathbf{x}-\mu_{3D}(t)$ and $\mathbf{x}$ denotes a 3D point.

Rendering then follows differentiable Gaussian splatting \cite{kerbl20233d}. Let $\alpha_i(t)$ be the projected opacity of the $i$-th sliced Gaussian at time $t$, and let $c_i(t)$ denote its RGB color evaluated from $c$ given the viewing direction. Pixel colors are computed by standard front-to-back alpha compositing:
\begin{equation}
C(t)=\sum_{i=1}^{N} c_i(t)\alpha_i(t) \prod_{j=1}^{i-1}(1-\alpha_j(t)),
\end{equation}
where Gaussians are ordered by depth along the ray.

\begin{table*}[t]
\centering
\renewcommand{\arraystretch}{0.82}
\setlength{\tabcolsep}{2.0pt}
\scriptsize
\resizebox{\textwidth}{!}{%
\begin{tabular}{l|cccc|cccc|cccc}
\toprule
\textbf{Method} &
\multicolumn{4}{c|}{\textbf{N3DV}} &
\multicolumn{4}{c|}{\textbf{Technicolor}} &
\multicolumn{4}{c}{\textbf{ENeRF-Outdoor}} \\
& PSNR$\uparrow$ & SSIM$_1$\,$\uparrow$ & SSIM$_2$\,$\uparrow$ & LPIPS$\downarrow$
& PSNR$\uparrow$ & SSIM$_1$\,$\uparrow$ & SSIM$_2$\,$\uparrow$ & LPIPS$\downarrow$
& PSNR$\uparrow$ & SSIM$_1$\,$\uparrow$ & SSIM$_2$\,$\uparrow$ & LPIPS$\downarrow$ \\
\midrule
\multicolumn{13}{c}{\textbf{2 Views}} \\
\midrule
4DGS & \capsecond{19.3465} & \capsecond{0.7251} & 0.7249 & \capthird{0.2295} & 13.8565 & \capthird{0.4919} & 0.4588 & \capthird{0.4447} & \capthird{16.1750} & \capthird{0.3120} & 0.3348 & 0.4528 \\
Ex4DGS & 17.2522 & 0.6622 & 0.7440 & 0.2932 & 14.0678 & 0.4191 & 0.5479 & 0.4662 & 13.9378 & 0.2496 & 0.4133 & 0.5156 \\
CEM-4DGS & 17.8677 & 0.6896 & \capthird{0.7657} & 0.2717 & 14.3640 & 0.4267 & 0.5546 & 0.4595 & 13.7995 & 0.2575 & \capthird{0.4264} & 0.5086 \\
STGS & 18.5025 & 0.7029 & 0.6825 & 0.2577 & \capsecond{20.7975} & \capsecond{0.7613} & \capsecond{0.7508} & \capsecond{0.2050} & 15.9265 & \capsecond{0.3515} & 0.3295 & \capbest{0.3555} \\
Swift4D & 15.5046 & 0.6549 & 0.6360 & 0.3478 & 14.1178 & 0.4068 & 0.3293 & 0.4999 & 15.8813 & 0.3043 & 0.3883 & \capthird{0.4060} \\
4C4D & \capthird{19.2770} & \capthird{0.7047} & \capsecond{0.7936} & \capsecond{0.2171} & \capthird{14.4915} & 0.4149 & \capthird{0.5857} & 0.5295 & \capsecond{16.2983} & 0.3023 & \capbest{0.5306} & 0.4299 \\
\textbf{D$^2$-4DGS} & \capbest{21.4668} & \capbest{0.8011} & \capbest{0.8189} & \capbest{0.1968} & \capbest{21.8570} & \capbest{0.7933} & \capbest{0.7853} & \capbest{0.1818} & \capbest{17.5073} & \capbest{0.4059} & \capsecond{0.4965} & \capsecond{0.3800} \\
\midrule
\multicolumn{13}{c}{\textbf{3 Views}} \\
\midrule
4DGS & \capthird{21.6551} & \capthird{0.7988} & 0.8305 & \capthird{0.1604} & 17.7284 & 0.5907 & 0.5838 & 0.3720 & \capsecond{19.5915} & 0.3658 & 0.4899 & 0.3432 \\
Ex4DGS & 19.8208 & 0.7603 & 0.8211 & 0.1797 & 24.0245 & 0.8002 & \capsecond{0.8730} & 0.1550 & 16.5998 & 0.3049 & 0.4788 & 0.4324 \\
CEM-4DGS & 21.6185 & 0.7951 & \capsecond{0.8500} & \capsecond{0.1486} & \capthird{24.2131} & \capthird{0.8046} & \capbest{0.8759} & \capthird{0.1472} & 16.4353 & 0.3052 & 0.4841 & 0.4177 \\
STGS & \capsecond{21.8865} & 0.7852 & 0.7728 & 0.1691 & \capsecond{26.0144} & \capsecond{0.8441} & 0.8374 & \capsecond{0.1277} & 18.9614 & \capsecond{0.3989} & 0.3806 & \capthird{0.2789} \\
Swift4D & 21.3068 & \capsecond{0.7996} & 0.8276 & 0.1692 & 14.9963 & 0.4529 & 0.4090 & 0.4546 & \capthird{19.3271} & \capthird{0.3880} & \capthird{0.5768} & \capbest{0.2582} \\
4C4D & 21.4542 & 0.7642 & \capthird{0.8420} & 0.1640 & 16.9738 & 0.5177 & 0.6764 & 0.4563 & 18.9710 & 0.3678 & \capsecond{0.5862} & 0.3413 \\
\textbf{D$^2$-4DGS} & \capbest{22.7561} & \capbest{0.8349} & \capbest{0.8589} & \capbest{0.1467} & \capbest{27.3255} & \capbest{0.8684} & \capthird{0.8639} & \capbest{0.1098} & \capbest{20.7804} & \capbest{0.4244} & \capbest{0.6125} & \capsecond{0.2696} \\
\midrule
\multicolumn{13}{c}{\textbf{4 Views}} \\
\midrule
4DGS & \capsecond{23.3693} & \capsecond{0.8411} & 0.8705 & \capthird{0.1253} & 19.3320 & 0.6070 & 0.5952 & 0.3648 & \capsecond{20.7969} & \capthird{0.4116} & 0.5980 & 0.3143 \\
Ex4DGS & 22.0370 & 0.8185 & 0.8690 & 0.1345 & \capthird{25.5925} & 0.8103 & \capsecond{0.8826} & 0.1479 & 17.7221 & 0.3248 & 0.5039 & 0.3997 \\
CEM-4DGS & 22.5162 & 0.8303 & \capthird{0.8782} & \capsecond{0.1216} & 25.5847 & \capthird{0.8142} & \capbest{0.8849} & \capthird{0.1395} & 18.1062 & 0.3287 & 0.5104 & 0.3789 \\
STGS & 22.8793 & 0.8249 & 0.8187 & 0.1320 & \capsecond{27.7342} & \capsecond{0.8514} & 0.8448 & \capsecond{0.1197} & 20.2357 & \capsecond{0.4169} & 0.3958 & \capthird{0.2604} \\
Swift4D & 22.3474 & \capthird{0.8354} & 0.8695 & 0.1254 & 16.1208 & 0.5035 & 0.4741 & 0.4109 & \capthird{20.7893} & 0.4027 & \capthird{0.6028} & \capbest{0.2429} \\
4C4D & \capthird{23.0567} & 0.8198 & \capsecond{0.8840} & 0.1280 & 18.5855 & 0.5519 & 0.7075 & 0.4450 & 20.5427 & 0.3927 & \capsecond{0.6091} & 0.3117 \\
\textbf{D$^2$-4DGS} & \capbest{24.9534} & \capbest{0.8636} & \capbest{0.9000} & \capbest{0.1211} & \capbest{29.1182} & \capbest{0.8763} & \capthird{0.8720} & \capbest{0.1061} & \capbest{22.0253} & \capbest{0.5205} & \capbest{0.6500} & \capsecond{0.2557} \\
\bottomrule
\end{tabular}%
}
\caption{Quantitative comparisons with state-of-the-art dynamic Gaussian Splatting methods on N3DV, Technicolor, and ENeRF-Outdoor under 2/3/4-view sparse-camera settings. We report PSNR, SSIM$_1$, SSIM$_2$, and LPIPS, where higher PSNR/SSIM and lower LPIPS indicate better performance. Results are marked as \legendbest{best score}, \legendsecond{second best score}, and \legendthird{third best score}.}
\label{tab:main_results}
\end{table*}

\begin{figure*}[t]
    \centering
    \includegraphics[width=\textwidth]{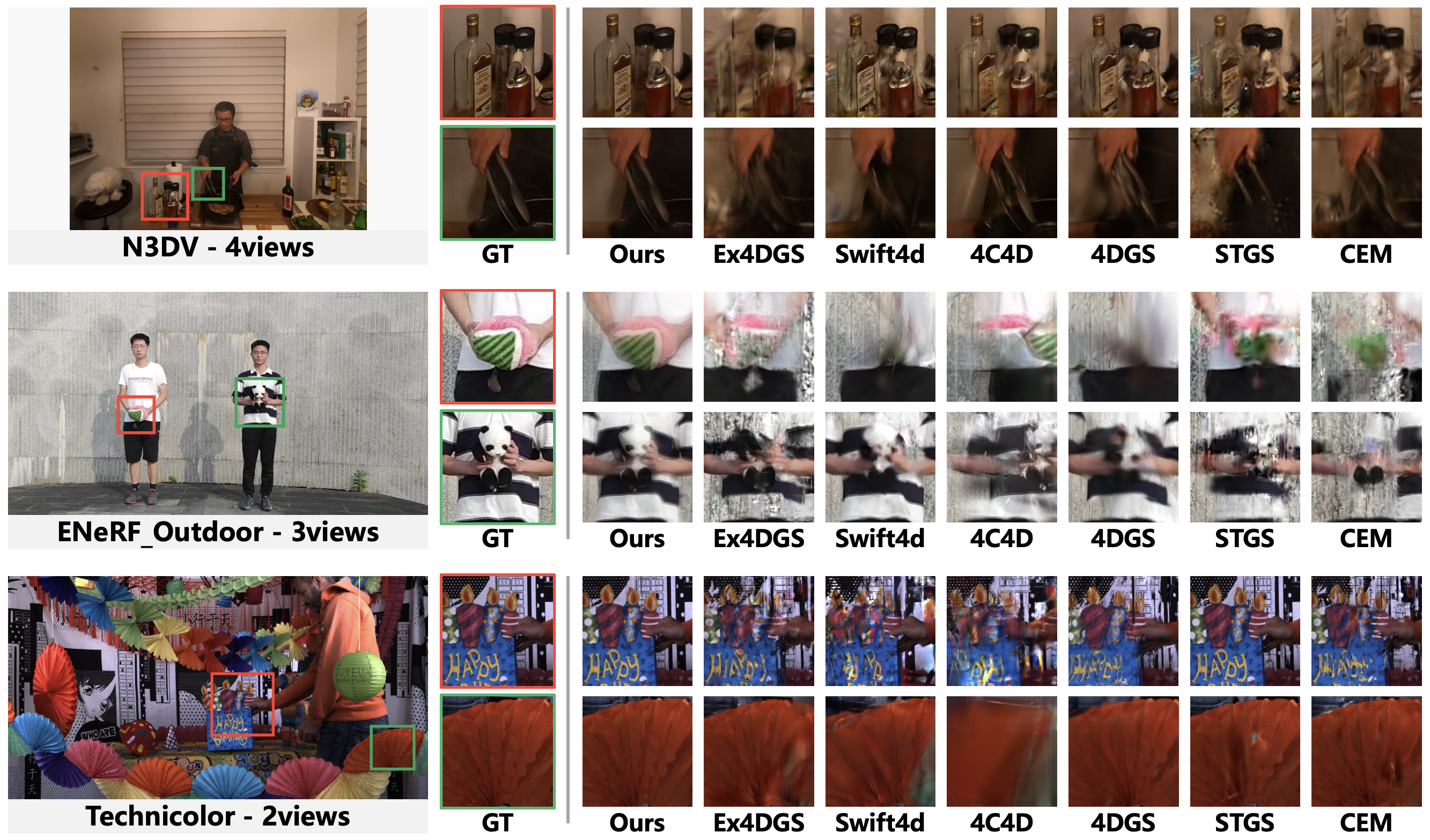}
    \caption{Qualitative comparisons on N3DV with 4 views, ENeRF-Outdoor with 3 views, and Technicolor with 2 views. Two enlarged regions are shown for each scene. Compared with existing methods, D$^2$-4DGS reconstructs sharper object boundaries and more coherent fine details while reducing foreground-background blending, structural distortion, and floating artifacts under sparse-view supervision.}
    \label{fig:qualitative_compare}
\end{figure*}

\begin{figure*}[t]
    \centering
    \includegraphics[width=\textwidth]{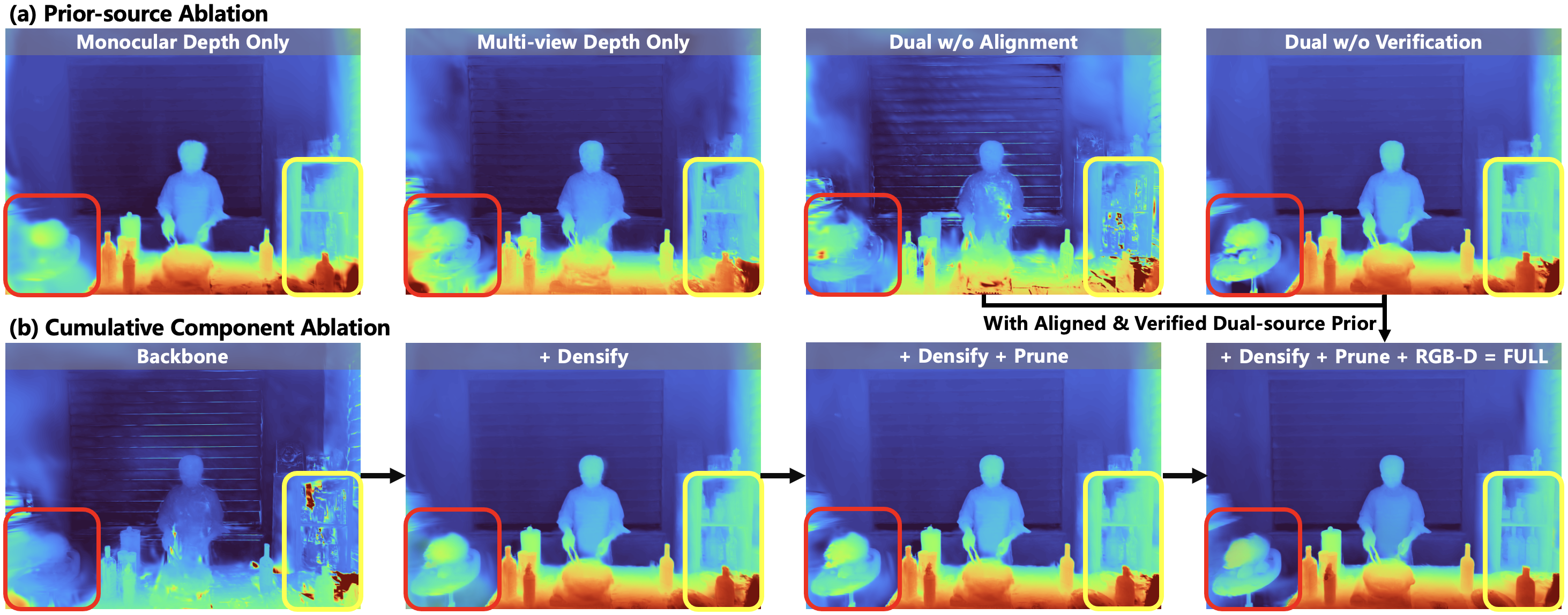}
    \caption{Depth visualizations for ablation on the \emph{sear\_steak} scene of N3DV under the 4-view setting. (a) We compare mono-only, MVS-only, unaligned dual-source, and unverified dual-source priors using identical densification, pruning, and RGB-D optimization. (b) With alignment and cross-source verification enabled, we cumulatively add depth-guided densification, pruning, and RGB-D optimization to the backbone. Colored boxes highlight progressively recovered surfaces and reduced local geometric inconsistencies.}
    \label{fig:ablation_depth}
\end{figure*}

\section{Methodology}

Given synchronized multi-view RGB videos
$\mathcal{I}=\{I_{t,n}\mid t\in\mathcal{T},n\in\mathcal{V}\}$
from a sparse set of calibrated cameras, we reconstruct a dynamic 4D Gaussian representation for novel view synthesis from limited observations. As shown in Fig.~\ref{fig:pipeline}, our implementation operates on rendered RGB/depth maps and Gaussian primitives while leaving the selected temporal parameterization unchanged. Here $t$, $n$, and $\mathbf{p}$ denote a timestamp, camera index, and pixel coordinate, respectively; $\epsilon$ ensures numerical stability.

D$^2$-4DGS couples depth-prior construction with Gaussian structure optimization. Aligned monocular and multi-view geometric depths identify verified anchors, while mono-only depths remain densification proposals (Sec.~\ref{sec:dual_depth_prior}). These cues guide Gaussian insertion, whereas pruning combines verified depth consistency with visibility-aware opacity contribution (Sec.~\ref{sec:densify_prune}). RGB reconstruction and verified depth supervision then refine the updated representation (Sec.~\ref{sec:rgbd_optimization}).

\subsection{Dual-source Depth Consistency Prior Construction}
\label{sec:dual_depth_prior}

We obtain monocular depth $D^m_{t,n}$ independently for each frame using Depth Anything V2 \cite{yang2024depth} and multi-view geometric depth $D^g_{t,n}$ using COLMAP PatchMatch Stereo \cite{schonberger2016pixelwise} across synchronized calibrated views at timestamp $t$. No frames from other timestamps are used. COLMAP geometric-consistency filtering removes unreliable multi-view estimates, while cross-source verification discards conflicting predictions where both depths are available. Monocular depth provides dense structure but is scale-ambiguous, whereas geometric depth is consistent with the reconstruction coordinate system but incomplete under sparse views, occlusion, and weak texture.

Let $M^g_{t,n}$ retain depths passing COLMAP's geometric-consistency check and $\Omega^g_{t,n}=\{\mathbf{p}\mid M^g_{t,n}(\mathbf{p})=1\}$. For each frame and view, we independently fit a scale and shift, constraining the scale to be positive, using the Charbonnier penalty $\rho(x)=\sqrt{x^2+\epsilon_\rho^2}-\epsilon_\rho$:
\begin{equation}
(a_{t,n},b_{t,n})=
\operatorname*{arg\,min}_{a>0,b}
\sum_{\mathbf{p}\in\Omega^g_{t,n}}
\rho\!\left(aD^m_{t,n}(\mathbf{p})+b-D^g_{t,n}(\mathbf{p})\right).
\end{equation}
The aligned monocular depth is
\begin{equation}
\bar D^m_{t,n}(\mathbf{p})=a_{t,n}D^m_{t,n}(\mathbf{p})+b_{t,n}.
\end{equation}
We measure cross-source disagreement and define the verified region as
\begin{equation}
\begin{aligned}
e_{t,n}(\mathbf{p})
&=\frac{|\bar D^m_{t,n}(\mathbf{p})-D^g_{t,n}(\mathbf{p})|}
{D^g_{t,n}(\mathbf{p})+\epsilon},\\
\Omega^v_{t,n}
&=\{\mathbf{p}\in\Omega^g_{t,n}\mid e_{t,n}(\mathbf{p})<\tau_c\}.
\end{aligned}
\end{equation}
The relative discrepancy reduces sensitivity to scene scale. The dual-source prior is
\begin{equation}
D^\star_{t,n}(\mathbf{p})=
\begin{cases}
D^g_{t,n}(\mathbf{p}), & \mathbf{p}\in\Omega^v_{t,n},\\
\bar D^m_{t,n}(\mathbf{p}), & M^g_{t,n}(\mathbf{p})=0 \land \bar D^m_{t,n}(\mathbf{p})>0,\\
\emptyset, & \mathrm{otherwise}.
\end{cases}
\end{equation}
Its support is $\Omega^p_{t,n}=\{\mathbf{p}\mid D^\star_{t,n}(\mathbf{p})\neq\emptyset\}$. Verified pixels in $\Omega^v_{t,n}$ support pruning and depth supervision, mono-only pixels in $\Omega^p_{t,n}\setminus\Omega^v_{t,n}$ only propose densification, and conflicting pixels are discarded.

\subsection{Dual-source Depth Prior-guided 4D Gaussian Densification and Pruning}
\label{sec:densify_prune}

The prior directly updates Gaussian structure rather than serving only as a rendering loss. Under sparse-view RGB supervision, visible surfaces can remain under-reconstructed, while floating primitives may still explain the training views. We use prior support to locate missing geometry and verified anchors to evaluate existing primitives.

For Gaussians contributing to pixel $\mathbf{p}$ in front-to-back order, let $\alpha_i(\mathbf{p})$ denote the projected opacity of Gaussian $i$. Its transmittance and effective compositing weight are $T_i(\mathbf{p})=\prod_{j<i}(1-\alpha_j(\mathbf{p}))$ and $w_i(\mathbf{p})=T_i(\mathbf{p})\alpha_i(\mathbf{p})$. Let $z_i(\mathbf{p})$ denote its view-space depth and $A_{t,n}(\mathbf{p})=\sum_{i\in\mathcal R(\mathbf{p})}w_i(\mathbf{p})$. We render depth and determine its validity by
\begin{equation}
\begin{aligned}
\hat D_{t,n}(\mathbf{p})
&=\frac{\sum_{i\in\mathcal R(\mathbf{p})}w_i(\mathbf{p})z_i(\mathbf{p})}
{A_{t,n}(\mathbf{p})+\epsilon},\\
M^r_{t,n}(\mathbf{p})
&=\mathbf{1}[A_{t,n}(\mathbf{p})>\tau_v].
\end{aligned}
\end{equation}
Thus, $M^r_{t,n}(\mathbf{p})=0$ indicates insufficient contribution for reliable surface depth. Where rendered depth is valid, its relative discrepancy is
\begin{equation}
\delta_{t,n}(\mathbf{p})
=
\frac{|\hat D_{t,n}(\mathbf{p})-D^\star_{t,n}(\mathbf{p})|}
{D^\star_{t,n}(\mathbf{p})+\epsilon}.
\end{equation}
The under-reconstructed region is
\begin{equation}
\hspace{-0.6em}
\mathcal U_{t,n}
=
\left\{
\mathbf{p}\in\Omega^p_{t,n}
\;\middle|\;
\begin{gathered}
M^r_{t,n}(\mathbf{p})=0
\;\mathrm{or}\\
M^r_{t,n}(\mathbf{p})=1
\land
\delta_{t,n}(\mathbf{p})>\tau_u
\end{gathered}
\right\}.
\end{equation}
The two conditions detect missing surface responses and inconsistent rendered depths, respectively. Sampled pixels are back-projected as
\begin{equation}
\mathbf{x}_{t,n,\mathbf{p}}=
\Pi_n^{-1}(\mathbf{p},D^\star_{t,n}(\mathbf{p})),\qquad
\mathbf{p}\in\mathcal U_{t,n}.
\end{equation}
New Gaussians are initialized at these points with corresponding pixel colors; other attributes follow the selected dynamic Gaussian initialization. Sampling and initialization details are provided in the supplementary material.

For pruning, let $\mathcal Q^v_k$ contain verified pixels to which Gaussian $g_k$ contributes, $\mathcal V_k$ its visible observations, and $\mathcal P_{k,t,n}$ its projected footprint. Its normalized depth inconsistency is
\begin{equation}
\begin{aligned}
\widetilde E_k&=
\frac{\sum_{q\in\mathcal Q^v_k}w_{k,q}
\frac{|z_{k,q}-D^\star_q|}{D^\star_q+\epsilon}}
{\sum_{q\in\mathcal Q^v_k}w_{k,q}+\epsilon},\\
E_k&=\min\!\left(\frac{\widetilde E_k}{\tau_u},1\right),
\end{aligned}
\end{equation}
where $q=(t,n,\mathbf{p})$. The visibility-aware opacity contribution and pruning score are
\begin{equation}
\begin{aligned}
O_k&=\frac{1}{|\mathcal V_k|}\sum_{(t,n)\in\mathcal V_k}
\max_{\mathbf{p}\in\mathcal P_{k,t,n}}w_{k,t,n}(\mathbf{p}),\\
S_k&=\lambda_eE_k+\lambda_o(1-O_k),
\quad \lambda_e,\lambda_o\geq0,\quad
\lambda_e+\lambda_o=1.
\end{aligned}
\end{equation}
Both signals lie in $[0,1]$. A large $E_k$ indicates geometric inconsistency, while a small $O_k$ indicates weak opacity contribution after accounting for visibility and occlusion. We update the pruning state of $g_k$ only when it contributes to verified-depth pixels, and remove it only when $S_k>\tau_p$ persists across consecutive such evaluations.

\subsection{Dual-source Depth-constrained 4D Gaussian RGB-D Joint Optimization}
\label{sec:rgbd_optimization}

RGB-D refinement alternates with scheduled structure updates and continues after densification and pruning stop. The rendered depth above is differentiable with respect to contributing Gaussians and is constrained only where verified prior depth and valid rendered depth coexist. Let $r^D_{t,n}(\mathbf{p})=(\hat D_{t,n}(\mathbf{p})-D^\star_{t,n}(\mathbf{p}))/(D^\star_{t,n}(\mathbf{p})+\epsilon)$. The final objective is
\begin{equation}
\begin{aligned}
\mathcal L&=\mathcal L_{rgb}+\lambda_d\mathcal L_{depth},\\
\mathcal L_{depth}&=
\frac{\sum_{\mathbf{p}\in\Omega^v_{t,n}}M^r_{t,n}(\mathbf{p})
\rho\!\left(r^D_{t,n}(\mathbf{p})\right)}
{\sum_{\mathbf{p}\in\Omega^v_{t,n}}M^r_{t,n}(\mathbf{p})+\epsilon}.
\end{aligned}
\end{equation}
Because it combines cross-source verification with rendered-depth validity, the mask excludes conflicting priors and pixels without a stable surface response.
Thus, mono-only regions propose structural completion without becoming direct constraints, verified depth improves geometry, and RGB supervision preserves appearance. The exact photometric objective and shared method settings are provided in the supplementary material.

\begin{figure}[t]
    \centering
    \includegraphics[width=\columnwidth]{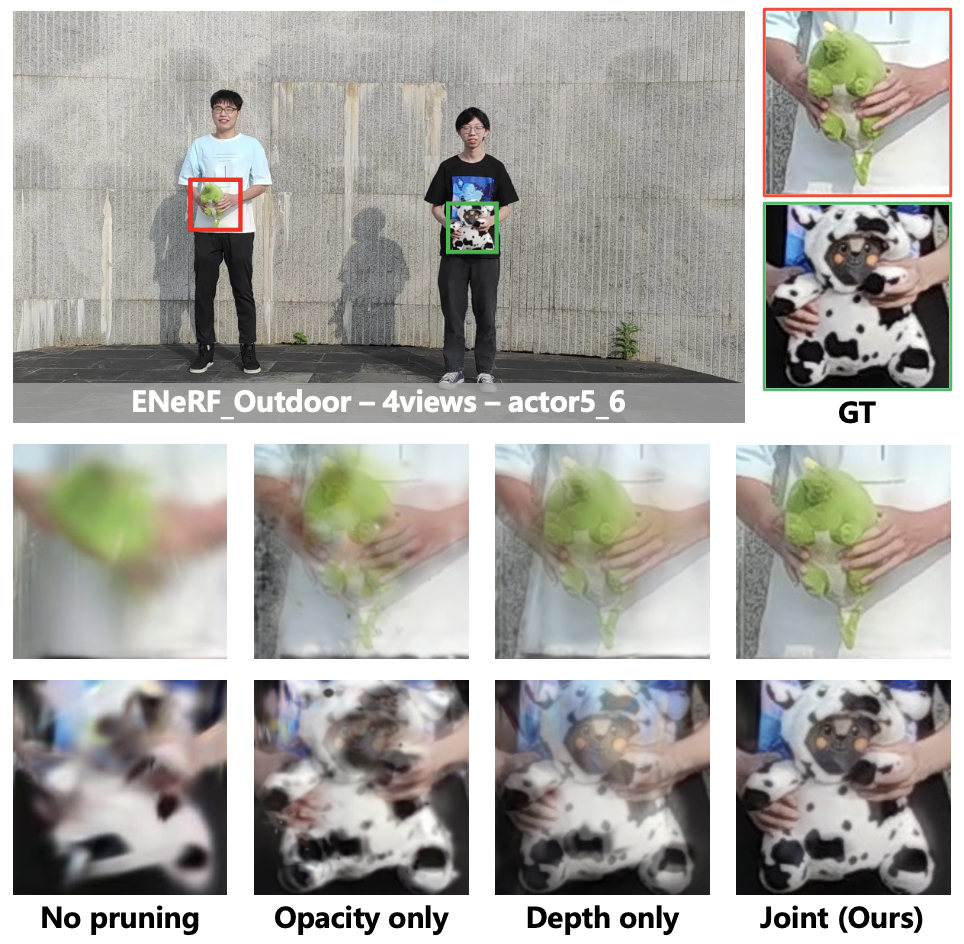}
    \caption{Visual analysis of pruning signals on
    \emph{actor5\_6} from ENeRF-Outdoor under the 4-view setting.
    The enlarged regions compare no pruning, opacity-only pruning,
    depth-only pruning, and their joint use. Opacity-based pruning
    removes weakly contributing primitives but may retain geometrically
    incorrect Gaussians, whereas depth-based pruning better suppresses
    inconsistent geometry. Combining both signals produces cleaner
    object boundaries and more coherent local details.}
    \label{fig:pruning_analysis}
\end{figure}

\section{Experiments}

\subsection{Experimental Settings}

\textbf{Datasets.}
We evaluate D$^2$-4DGS on six N3DV scenes \cite{li2022neural}, five Technicolor scenes \cite{sabater2017dataset}, and three ENeRF-Outdoor scenes \cite{lin2022efficient} under 2/3/4-view settings. Only designated cameras are used for training, while all other valid cameras are held out for evaluation. Scene lists, image resolutions, frame counts, and camera splits are provided in the supplementary material.

\textbf{Baselines.}
We compare D$^2$-4DGS with 4DGaussians \cite{wu20244d}, Ex4DGS \cite{lee2024fully}, CEM-4DGS \cite{kang2025clustered}, Spacetime Gaussians \cite{li2024spacetime}, Swift4D \cite{wu2025swift4d}, and 4C4D \cite{zhou20264c4d}. All methods use identical training-camera subsets and held-out test views. Although designed for four cameras, 4C4D accepts arbitrary training-view lists; we therefore evaluate it on the same 2/3/4-view subsets, retaining its official settings and regenerating initialization for each subset.

\textbf{Metrics.}
We report PSNR, SSIM$_1$, SSIM$_2$, and LPIPS averaged over test views and timestamps. SSIM$_1$ and SSIM$_2$ use data ranges of 1.0 and 2.0, respectively, while lower LPIPS is better.

\textbf{Implementation Details.}
All experiments are conducted on a single NVIDIA RTX 4090 GPU with 24GB memory. Verified anchors guide pruning and depth supervision in RGB-D optimization; verified geometric depths and aligned mono-only estimates support densification. Unless otherwise specified, hyperparameters are shared across datasets and view settings. Dataset splits and additional implementation details are provided in the supplementary material.

\begin{table}[t]
    \centering
    \renewcommand{\arraystretch}{0.96}
    \setlength{\tabcolsep}{2.4pt}

    {\footnotesize
    \resizebox{\columnwidth}{!}{%
    \begin{tabular}{@{}llcccc@{}}
        \toprule
        \textbf{Study} & \textbf{Variant}
        & \textbf{PSNR}$\uparrow$
        & \textbf{SSIM$_1$}$\uparrow$
        & \textbf{SSIM$_2$}$\uparrow$
        & \textbf{LPIPS}$\downarrow$ \\
        \midrule

        \textsc{Base}
        & Backbone
        & 25.3987 & 0.8784 & 0.9067 & 0.1136 \\

        \midrule
        \textsc{Prior}
        & Mono only
        & 25.9236 & 0.8840 & 0.9145 & 0.1098 \\
        & MVS only
        & 25.6174 & 0.8815 & 0.9106 & 0.1115 \\
        & Dual w/o align.
        & 25.8469 & 0.8832 & 0.9130 & 0.1107 \\
        & Dual w/o verify.
        & 26.1183 & 0.8875 & 0.9188 & 0.1088 \\

        \midrule
        \textsc{Component}
        & w/o Densification$^\dagger$
        & 26.0384 & 0.8849 & 0.9161 & 0.1096 \\
        & w/o Pruning
        & 26.5217 & 0.8915 & 0.9231 & 0.1059 \\
        & w/o RGB-D$^\ddagger$
        & 26.4472 & 0.8910 & 0.9225 & 0.1068 \\

        \midrule
        \textsc{Full}
        & \textbf{D$^2$-4DGS}
        & \textbf{27.2302}
        & \textbf{0.9023}
        & \textbf{0.9337}
        & \textbf{0.0924} \\
        \bottomrule
    \end{tabular}%
    }

    \parbox{\columnwidth}{\footnotesize
        $^\dagger$ Uses the backbone's original densification strategy;
        $^\ddagger$ retains RGB supervision without verified-depth regularization.}
    }

    \caption{Ablation study on the \emph{sear\_steak} scene of N3DV
    under the 4-view setting. Prior variants use identical densification,
    pruning, and RGB-D optimization, while component variants ablate one
    proposed operation. Higher PSNR/SSIM and lower LPIPS indicate better
    performance.}
    \label{tab:ablation}
\end{table}

\subsection{Results and Evaluation}

\textbf{Quantitative Comparisons.}
Table~\ref{tab:main_results} reports rendering quality on held-out views under 2/3/4-view settings. D$^2$-4DGS delivers leading PSNR together with strong structural and perceptual scores across datasets and view settings, indicating that dual-source depth guidance improves fidelity while maintaining competitive structural and perceptual quality. The gains are particularly clear with two views, where weak cross-view constraints make RGB-driven Gaussian placement ambiguous and let unstable primitives fit the training observations. PSNR gains persist with more views, confirming that the method remains robust across different sparsity levels and diverse scene characteristics.

\textbf{Qualitative Comparisons.}
Figure~\ref{fig:qualitative_compare} compares N3DV with 4 views, ENeRF-Outdoor with 3 views, and Technicolor with 2 views. Existing methods blur object boundaries and hand-object interactions, blend foreground objects with the background, or lose high-frequency decorations. These failures occur because geometrically incorrect Gaussians can still fit the sparse training observations. Using verified anchors for pruning and depth supervision, together with verified geometric depths and aligned mono-only estimates for densification, D$^2$-4DGS reconstructs more coherent foreground structures, sharper boundaries, and finer details across scenes and sparsity levels.

\subsection{Ablation Study}
\label{sec:ablation}

We conduct ablations on \emph{sear\_steak} from N3DV under identical 4-view settings. Table~\ref{tab:ablation} compares four depth-prior variants using identical densification, pruning, and RGB-D optimization, then ablates each proposed operation. The densification variant uses the backbone strategy, while \emph{w/o RGB-D} retains RGB supervision without verified-depth regularization.

Single-source priors improve the backbone, confirming the benefit of depth guidance. Mono-only outperforms MVS-only, reflecting dense monocular coverage when geometric depth is incomplete. Neither single-source nor unaligned or unverified dual-source guidance reaches the full model, showing reliable fusion requires alignment and verification. Each component ablation degrades performance, with densification causing the largest decline, highlighting its importance for recovering missing surfaces. Pruning suppresses unstable primitives, while verified-depth regularization refines geometry. Figure~\ref{fig:ablation_depth} shows their complementarity, as cumulative additions recover incomplete regions, reduce local inconsistencies, and produce cleaner boundaries.

\subsection{Analysis of Pruning Signals}
\label{sec:pruning_analysis}

We analyze the pruning signals on \emph{actor5\_6} from ENeRF-Outdoor under the 4-view setting while keeping the dual-source prior, densification, RGB-D objective, and training schedule fixed and varying only the pruning criterion. Table~\ref{tab:pruning_criteria} shows that depth inconsistency is more informative than visibility-aware opacity contribution alone, while their combination achieves the best results across all four reported metrics. The \emph{Opacity only} variant removes weakly contributing primitives but may retain strongly contributing Gaussians at incorrect depths, whereas depth inconsistency directly captures such geometric errors. Figure~\ref{fig:pruning_analysis} further demonstrates their complementarity, with the joint criterion preserving object details while producing cleaner boundaries and fewer floating artifacts.

\begin{table}[t]
    \centering
    \renewcommand{\arraystretch}{0.94}
    \setlength{\tabcolsep}{1.8pt}

    {\footnotesize
    \begin{tabular*}{\columnwidth}
    {@{\extracolsep{\fill}}lcccc@{}}
        \toprule
        \textbf{Criterion}
        & \textbf{PSNR}$\uparrow$
        & \textbf{SSIM$_1$}$\uparrow$
        & \textbf{SSIM$_2$}$\uparrow$
        & \textbf{LPIPS}$\downarrow$ \\
        \midrule

        No pruning
        & 21.1874 & 0.4618 & 0.5662 & 0.2549 \\

        Opacity only
        & 21.5663 & 0.4789 & 0.6107 & 0.2534 \\

        Depth only
        & 21.9076 & 0.5015 & 0.6428 & 0.2514 \\

        \textbf{Joint (Ours)}
        & \textbf{22.2082}
        & \textbf{0.5183}
        & \textbf{0.6693}
        & \textbf{0.2496} \\
        \bottomrule
    \end{tabular*}
    }

    \caption{Analysis of pruning signals on
    \emph{actor5\_6} from ENeRF-Outdoor under the 4-view setting.}
    \label{tab:pruning_criteria}
\end{table}

\section{Conclusion}

We presented D$^2$-4DGS, a dual-depth-guided framework for dynamic novel view synthesis from a sparse set of synchronized cameras. Without introducing an additional temporal motion model, D$^2$-4DGS strengthens the geometric guidance available to dynamic Gaussian optimization. It aligns complementary monocular and multi-view depths, using verified anchors for pruning and depth supervision while retaining verified geometric depths and aligned mono-only estimates as densification proposals. RGB-D joint optimization then refines appearance and geometry. Experiments on N3DV, Technicolor, and ENeRF-Outdoor under 2/3/4-view settings demonstrate consistent PSNR gains with competitive structural and perceptual quality, showing that reliable depth-guided structure optimization can effectively alleviate geometric ambiguity with sparse-camera inputs.

\textbf{Limitations and future work.}
Dual-depth preprocessing and structure updates increase training time. Where multi-view depth is unavailable, densification relies on aligned monocular estimates, and large errors may reduce structural-completion accuracy. Future work will explore lightweight depth priors and efficient update schedules.

\bibliography{aaai2027}

@article{kerbl20233d,
  title={3d gaussian splatting for real-time radiance field rendering.},
  author={Kerbl, Bernhard and Kopanas, Georgios and Leimk{\"u}hler, Thomas and Drettakis, George and others},
  journal={ACM Trans. Graph.},
  volume={42},
  number={4},
  pages={139--1},
  year={2023}
}

@inproceedings{wu20244d,
  title={4d gaussian splatting for real-time dynamic scene rendering},
  author={Wu, Guanjun and Yi, Taoran and Fang, Jiemin and Xie, Lingxi and Zhang, Xiaopeng and Wei, Wei and Liu, Wenyu and Tian, Qi and Wang, Xinggang},
  booktitle={Proceedings of the IEEE/CVF conference on computer vision and pattern recognition},
  pages={20310--20320},
  year={2024}
}

@inproceedings{li2024spacetime,
  title={Spacetime gaussian feature splatting for real-time dynamic view synthesis},
  author={Li, Zhan and Chen, Zhang and Li, Zhong and Xu, Yi},
  booktitle={Proceedings of the IEEE/CVF Conference on Computer Vision and Pattern Recognition},
  pages={8508--8520},
  year={2024}
}

@article{yang2024depth,
  title={Depth anything v2},
  author={Yang, Lihe and Kang, Bingyi and Huang, Zilong and Zhao, Zhen and Xu, Xiaogang and Feng, Jiashi and Zhao, Hengshuang},
  journal={Advances in Neural Information Processing Systems},
  volume={37},
  pages={21875--21911},
  year={2024}
}

@inproceedings{schonberger2016structure,
  title={Structure-from-motion revisited},
  author={Schonberger, Johannes L and Frahm, Jan-Michael},
  booktitle={Proceedings of the IEEE conference on computer vision and pattern recognition},
  pages={4104--4113},
  year={2016}
}

@inproceedings{schonberger2016pixelwise,
  title={Pixelwise view selection for unstructured multi-view stereo},
  author={Sch{\"o}nberger, Johannes L and Zheng, Enliang and Frahm, Jan-Michael and Pollefeys, Marc},
  booktitle={European conference on computer vision},
  pages={501--518},
  year={2016},
  organization={Springer}
}

@inproceedings{yang2024deformable,
  title={Deformable 3d gaussians for high-fidelity monocular dynamic scene reconstruction},
  author={Yang, Ziyi and Gao, Xinyu and Zhou, Wen and Jiao, Shaohui and Zhang, Yuqing and Jin, Xiaogang},
  booktitle={Proceedings of the IEEE/CVF conference on computer vision and pattern recognition},
  pages={20331--20341},
  year={2024}
}

@inproceedings{luiten2024dynamic,
  title={Dynamic 3d gaussians: Tracking by persistent dynamic view synthesis},
  author={Luiten, Jonathon and Kopanas, Georgios and Leibe, Bastian and Ramanan, Deva},
  booktitle={2024 International Conference on 3D Vision (3DV)},
  pages={800--809},
  year={2024},
  organization={IEEE}
}

@inproceedings{duan20244d,
  title={4d-rotor gaussian splatting: towards efficient novel view synthesis for dynamic scenes},
  author={Duan, Yuanxing and Wei, Fangyin and Dai, Qiyu and He, Yuhang and Chen, Wenzheng and Chen, Baoquan},
  booktitle={ACM SIGGRAPH 2024 Conference Papers},
  pages={1--11},
  year={2024}
}

@inproceedings{yang2024real,
  title={Real-time photorealistic dynamic scene representation and rendering with 4d gaussian splatting},
  author={Yang, Zeyu and Yang, Hongye and Pan, Zijie and Zhang, Li},
  booktitle={International Conference on Learning Representations},
  volume={2024},
  pages={9142--9159},
  year={2024}
}

@inproceedings{lin2024gaussian,
  title={Gaussian-flow: 4d reconstruction with dynamic 3d gaussian particle},
  author={Lin, Youtian and Dai, Zuozhuo and Zhu, Siyu and Yao, Yao},
  booktitle={Proceedings of the IEEE/CVF Conference on Computer Vision and Pattern Recognition},
  pages={21136--21145},
  year={2024}
}

@article{guo2024motion,
  title={Motion-aware 3d gaussian splatting for efficient dynamic scene reconstruction},
  author={Guo, Zhiyang and Zhou, Wengang and Li, Li and Wang, Min and Li, Houqiang},
  journal={IEEE Transactions on Circuits and Systems for Video Technology},
  volume={35},
  number={4},
  pages={3119--3133},
  year={2024},
  publisher={IEEE}
}

@article{xiong2023sparsegs,
  title={Sparsegs: Real-time 360 sparse view synthesis using gaussian splatting},
  author={Xiong, Haolin and Muttukuru, Sairisheek and Upadhyay, Rishi and Chari, Pradyumna and Kadambi, Achuta},
  journal={arXiv e-prints},
  pages={arXiv--2312},
  year={2023}
}

@inproceedings{zhu2024fsgs,
  title={Fsgs: Real-time few-shot view synthesis using gaussian splatting},
  author={Zhu, Zehao and Fan, Zhiwen and Jiang, Yifan and Wang, Zhangyang},
  booktitle={European conference on computer vision},
  pages={145--163},
  year={2024},
  organization={Springer}
}

@inproceedings{chung2024depth,
  title={Depth-regularized optimization for 3d gaussian splatting in few-shot images},
  author={Chung, Jaeyoung and Oh, Jeongtaek and Lee, Kyoung Mu},
  booktitle={Proceedings of the IEEE/CVF Conference on Computer Vision and Pattern Recognition},
  pages={811--820},
  year={2024}
}

@inproceedings{zhang2024cor,
  title={Cor-gs: sparse-view 3d gaussian splatting via co-regularization},
  author={Zhang, Jiawei and Li, Jiahe and Yu, Xiaohan and Huang, Lei and Gu, Lin and Zheng, Jin and Bai, Xiao},
  booktitle={European conference on computer vision},
  pages={335--352},
  year={2024},
  organization={Springer}
}

@inproceedings{li2024dngaussian,
  title={Dngaussian: Optimizing sparse-view 3d gaussian radiance fields with global-local depth normalization},
  author={Li, Jiahe and Zhang, Jiawei and Bai, Xiao and Zheng, Jin and Ning, Xin and Zhou, Jun and Gu, Lin},
  booktitle={Proceedings of the IEEE/CVF conference on computer vision and pattern recognition},
  pages={20775--20785},
  year={2024}
}

@inproceedings{charatan2024pixelsplat,
  title={pixelsplat: 3d gaussian splats from image pairs for scalable generalizable 3d reconstruction},
  author={Charatan, David and Li, Sizhe Lester and Tagliasacchi, Andrea and Sitzmann, Vincent},
  booktitle={Proceedings of the IEEE/CVF conference on computer vision and pattern recognition},
  pages={19457--19467},
  year={2024}
}

@inproceedings{chen2024mvsplat,
  title={Mvsplat: Efficient 3d gaussian splatting from sparse multi-view images},
  author={Chen, Yuedong and Xu, Haofei and Zheng, Chuanxia and Zhuang, Bohan and Pollefeys, Marc and Geiger, Andreas and Cham, Tat-Jen and Cai, Jianfei},
  booktitle={European conference on computer vision},
  pages={370--386},
  year={2024},
  organization={Springer}
}

@article{fan2024instantsplat,
  title={Instantsplat: Sparse-view gaussian splatting in seconds},
  author={Fan, Zhiwen and Cong, Wenyan and Wen, Kairun and Wang, Kevin and Zhang, Jian and Ding, Xinghao and Xu, Danfei and Ivanovic, Boris and Pavone, Marco and Pavlakos, Georgios and others},
  journal={arXiv preprint arXiv:2403.20309},
  year={2024}
}

@inproceedings{zhou20264c4d,
  title={4C4D: 4 Camera 4D Gaussian Splatting},
  author={Zhou, Junsheng and Yang, Zhifan and Han, Liang and Zhang, Wenyuan and Shi, Kanle and Xu, Shenkun and Liu, Yu-Shen},
  booktitle={Proceedings of the IEEE/CVF Conference on Computer Vision and Pattern Recognition},
  pages={11829--11839},
  year={2026}
}

@article{ranftl2020towards,
  title={Towards robust monocular depth estimation: Mixing datasets for zero-shot cross-dataset transfer},
  author={Ranftl, Ren{\'e} and Lasinger, Katrin and Hafner, David and Schindler, Konrad and Koltun, Vladlen},
  journal={IEEE transactions on pattern analysis and machine intelligence},
  volume={44},
  number={3},
  pages={1623--1637},
  year={2020},
  publisher={IEEE}
}

@inproceedings{ranftl2021vision,
  title={Vision transformers for dense prediction},
  author={Ranftl, Ren{\'e} and Bochkovskiy, Alexey and Koltun, Vladlen},
  booktitle={Proceedings of the IEEE/CVF international conference on computer vision},
  pages={12179--12188},
  year={2021}
}

@inproceedings{yao2018mvsnet,
  title={Mvsnet: Depth inference for unstructured multi-view stereo},
  author={Yao, Yao and Luo, Zixin and Li, Shiwei and Fang, Tian and Quan, Long},
  booktitle={Proceedings of the European conference on computer vision (ECCV)},
  pages={767--783},
  year={2018}
}

@inproceedings{gu2020cascade,
  title={Cascade cost volume for high-resolution multi-view stereo and stereo matching},
  author={Gu, Xiaodong and Fan, Zhiwen and Zhu, Siyu and Dai, Zuozhuo and Tan, Feitong and Tan, Ping},
  booktitle={Proceedings of the IEEE/CVF conference on computer vision and pattern recognition},
  pages={2495--2504},
  year={2020}
}

@inproceedings{wang2021patchmatchnet,
  title={Patchmatchnet: Learned multi-view patchmatch stereo},
  author={Wang, Fangjinhua and Galliani, Silvano and Vogel, Christoph and Speciale, Pablo and Pollefeys, Marc},
  booktitle={Proceedings of the IEEE/CVF conference on computer vision and pattern recognition},
  pages={14194--14203},
  year={2021}
}

@inproceedings{ding2022transmvsnet,
  title={Transmvsnet: Global context-aware multi-view stereo network with transformers},
  author={Ding, Yikang and Yuan, Wentao and Zhu, Qingtian and Zhang, Haotian and Liu, Xiangyue and Wang, Yuanjiang and Liu, Xiao},
  booktitle={Proceedings of the IEEE/CVF conference on computer vision and pattern recognition},
  pages={8585--8594},
  year={2022}
}

@article{cao2022mvsformer,
  title={Mvsformer: Multi-view stereo by learning robust image features and temperature-based depth},
  author={Cao, Chenjie and Ren, Xinlin and Fu, Yanwei},
  journal={arXiv preprint arXiv:2208.02541},
  year={2022}
}

@inproceedings{li2022neural,
  title={Neural 3d video synthesis from multi-view video},
  author={Li, Tianye and Slavcheva, Mira and Zollhoefer, Michael and Green, Simon and Lassner, Christoph and Kim, Changil and Schmidt, Tanner and Lovegrove, Steven and Goesele, Michael and Newcombe, Richard and others},
  booktitle={Proceedings of the IEEE/CVF conference on computer vision and pattern recognition},
  pages={5521--5531},
  year={2022}
}

@inproceedings{sabater2017dataset,
  title={Dataset and pipeline for multi-view light-field video},
  author={Sabater, Neus and Boisson, Guillaume and Vandame, Benoit and Kerbiriou, Paul and Babon, Frederic and Hog, Matthieu and Gendrot, Remy and Langlois, Tristan and Bureller, Olivier and Schubert, Arno and others},
  booktitle={Proceedings of the IEEE conference on computer vision and pattern recognition Workshops},
  pages={30--40},
  year={2017}
}

@inproceedings{lin2022efficient,
  title={Efficient neural radiance fields for interactive free-viewpoint video},
  author={Lin, Haotong and Peng, Sida and Xu, Zhen and Yan, Yunzhi and Shuai, Qing and Bao, Hujun and Zhou, Xiaowei},
  booktitle={SIGGRAPH Asia 2022 conference papers},
  pages={1--9},
  year={2022}
}

@article{lee2024fully,
  title={Fully explicit dynamic gaussian splatting},
  author={Lee, Junoh and Won, ChangYeon and Jung, Hyunjun and Bae, Inhwan and Jeon, Hae-Gon},
  journal={Advances in Neural Information Processing Systems},
  volume={37},
  pages={5384--5409},
  year={2024}
}

@article{wu2025swift4d,
  title={Swift4d: Adaptive divide-and-conquer gaussian splatting for compact and efficient reconstruction of dynamic scene},
  author={Wu, Jiahao and Peng, Rui and Wang, Zhiyan and Xiao, Lu and Tang, Luyang and Yan, Jinbo and Xiong, Kaiqiang and Wang, Ronggang},
  journal={arXiv preprint arXiv:2503.12307},
  year={2025}
}

@inproceedings{kang2025clustered,
  title={Clustered Error Correction with Grouped 4D Gaussian Splatting},
  author={Kang, Taeho and Park, Jaeyeon and Lee, Kyungjin and Lee, Youngki},
  booktitle={Proceedings of the SIGGRAPH Asia 2025 Conference Papers},
  pages={1--12},
  year={2025}
}


\end{document}